\documentclass{article}

\usepackage{microtype}
\usepackage{graphicx}
\usepackage{subfigure}
\usepackage{booktabs} 

\usepackage{hyperref}

\usepackage[accepted]{icml2024}

\usepackage{amsmath}
\usepackage{amssymb}
\usepackage{mathtools}
\usepackage{amsthm}

\usepackage{amsfonts}
\usepackage{makecell}
\usepackage{multirow}
\usepackage{float}
\usepackage{colortbl}
\usepackage[capitalize,noabbrev]{cleveref}
\theoremstyle{plain}

\theoremstyle{definition}

\theoremstyle{remark}

\usepackage[textsize=tiny]{todonotes}

\icmltitlerunning{SFC: Achieve Accurate Fast Convolution under Low-precision Arithmetic}

\begin{document}

\twocolumn[
\icmltitle{SFC: Achieve Accurate Fast Convolution under Low-precision Arithmetic}



\icmlsetsymbol{equal}{*}

\begin{icmlauthorlist}
\icmlauthor{Liulu He}{njuee}
\icmlauthor{Yufei Zhao}{njuee}
\icmlauthor{Rui Gao}{njuee}
\icmlauthor{Yuan Du}{njuee,njusz}
\icmlauthor{Li Du}{njuee,njusz}
\end{icmlauthorlist}

\icmlaffiliation{njuee}{School of Electronic Science and Engineering, Nanjing University, Nanjing 210023, China}
\icmlaffiliation{njusz}{Interdisciplinary Research Center for Future Intelligent Chips (Chip-X), Nanjing University, Suzhou 215163, China}
\icmlcorrespondingauthor{Yuan Du}{yuandu@nju.edu.cn}
\icmlcorrespondingauthor{Li Du}{ldu@nju.edu.cn}

\icmlkeywords{Machine Learning, ICML}

\vskip 0.3in
]



\printAffiliationsAndNotice{}  

\begin{abstract}
Fast convolution algorithms, including Winograd and FFT, can efficiently accelerate convolution operations in deep models. 
However, these algorithms depend on high-precision arithmetic to maintain inference accuracy, which conflicts with the model quantization. 
To resolve this conflict and further improve the efficiency of quantized convolution, we proposes SFC, a new algebra transform for fast convolution by extending the Discrete Fourier Transform (DFT) with symbolic computing, in which only additions are required to perform the transformation at specific transform points, avoiding the calculation of irrational number and reducing the requirement for precision.
Additionally, we enhance convolution efficiency by introducing correction terms to convert invalid circular convolution outputs of the Fourier method into effective ones. 
The numerical error analysis is presented for the first time in this type of work and proves that our algorithms can provide a 3.68× multiplication reduction for 3×3 convolution, while the Winograd algorithm only achieves a 2.25× reduction with similarly low numerical errors.
Experiments carried out on benchmarks and FPGA show that our new algorithms can further improve the computation efficiency of quantized models while maintaining accuracy, surpassing both the quantization-alone method and existing works on fast convolution quantization.
\end{abstract}

\section{Introduction}


Convolution operations are a crucial component of many deep learning models. Due to their intensive computation requirements, optimizing convolution calculations is key to improving model deployment efficiency.
Fast Convolution Algorithms and Quantization are two distinct approaches to mitigate the computational burden.
Fast convolution algorithms can reduce the arithmetic complexity by multiplying inputs and kernel weights in the transformation domain. A 3×3 convolution performed with Winograd F(2×2, 3×3) algorithm consumes just $\frac{1}{2.25}$ multiplications compared to direct computing.
Whereas, model quantization reduces the cost of a single arithmetic operation and data transmission by converting high-precision floating numbers to low-precision integers. An int8 multiply-and-accumulate operation consumes only $\frac{1}{16}$ of the energy compared to a fp32 operation.


However, when attempting to combine existing fast convolution algorithms and model quantization to further improve computational efficiency, the model accuracy could be severely degraded. This is due to the significant increase in numerical error when the two methods are used together. 
For example, Winograd, a well-known fast convolution algorithm for small filter sizes \citep{lavin2016fast}, has ill-conditioned transformation matrices (with high condition numbers), necessitating the use of high-precision arithmetic to avoid numerical issues \citep{barabasz2020error}.
Another renowned fast algorithm for convolution is the Fast Fourier transform (FFT). While its transformation is well-conditioned, its irrational coefficients can introduce significant rounding errors, particularly under low-precision representation. 
Number Theoretic Transform (NTT) can achieve precise computation for integer convolution, but it involves module operations and the transformations for inputs and filters would extend the bit-width to be equivalent to that of the outputs.  

At present, two research approaches have been developed to tackle the aforementioned problem. One approach involves customizing the quantization method specifically optimized for fast convolution algorithms \citep{li2021lowino,chikin2022channel,andri2022going,tianqi2023towards}. However, this approach struggles to maintain satisfactory accuracy under int8 quantization for faster algorithms such as Winograd F(4×4, 3×3). The other approach is to explore new fast algorithms that are better fit for quantization \citep{liu2020efficient,alam2022winograd}. Nevertheless, these emerging algorithms encounter challenges in achieving low arithmetic complexity. 
These facts show that simultaneously achieving low arithmetic complexity and low-precision quantization while maintaining model accuracy is a persistent challenge, which, to our knowledge, no existing work has overcome.

This paper aims to develop a new efficient fast convolution algorithm with high numerical accuracy, which is compatible with model quantization techniques. We note that all the complex coefficients in the Fourier Transform have constant modulus 1, so the quantization error can be reduced by bypassing the direct calculation on these coefficients. To achieve this, we introduce symbolic computing to avoid the involvement of irrational numbers, where the transformation process contains only additions. This advance adapts the Fourier transform to low-precision arithmetic. We also discovered that conventional Fourier convolution does not fully utilize its circular convolution outputs, which significantly affects its efficiency, so we add correction terms to convert these neglected outputs into valid results. Moreover, we investigate the error generation mechanism of convolution algorithms and compare the numerical errors of direct convolution, Winograd, and our algorithms. The key contributions can be summarized as follows:
\begin{itemize}
\item[1.] We formulate an efficient and quantization-friendly fast convolution algorithm extended from Fourier Convolution. We employ symbolic computing to perform Discret Fourier Transformation by just additions, and introduce correction terms to fully utilize its circular convolution outputs and further enhance its efficiency.

\item[2.] We analyse the numerical error of direct convolution and other fast convolution algorithms. Our method can achieve 3.68× arithmetic reduction while the Winogard algorithm only achieves 2.25× at equivalent numerical accuracy. Through error analysis and observation of the energy distribution in the frequency domain, we also present a frequency-wise quantization strategy to improve model accuracy at low-bitwidth.

\item[3.] Experiments on the ImageNet dataset validate that our method can achieve less than 0.2\% accuracy degradation with int8 post-training quantization. In the same model accuracy, our method achieves up to 2.5× bit-operations reduction compared to Winograd convolution or direct convolution under quantization. FPGA implementation shows that our algorithms can significantly improve model inference throughput combined with low-precision arithmetic.

\end{itemize}

\section{Related Work}
\textbf{Fast Convolution Algorithms.} The FFT was the first utilized algorithm \citep{mathieu2014fast} to fast the training of convolutions. Subsequently, for small convolutions, the Winograd minimum filtering algorithm \citep{lavin2016fast} was found to outperform the Fourier-based method due to its real field arithmetic operations, whereas the Fourier method requires more inefficient arithmetic in complex field. Additionally, the NTT has also been proposed to accelerate convolutions \citep{hong2022accelerating}. However, when combining quantization and fast convolution algorithms, the challenge of potential model accuracy degradation arises. The Winograd algorithm is susceptible to numerical instability due to the ill-conditioned Vandermonde matrix in the transformation \citep{vincent2017improving,barabasz2020error}. Fourier-based methods demand a high-precision format to accurately represent irrational numbers. NTT methods can offer accurate integer computing, but involve a large number of modulo operations and high-bitwidth intermediate data representations, reducing computation efficiency.

\textbf{Quantization for Fastconv.} Some approaches aim to optimize the quantization method to regain model accuracy. For example, LoWino \citep{li2021lowino} presents a post-training quantization (PTQ) method for Winograd, optimizing the scaling factor by minimizing the KL distance between the quantized and original vectors. Another PTQ work \citep{chikin2022channel} introduces a balancing operation between the filter and input channels to enhance bit-width utilization and improve the quality of quantization for Winograd. Additionally, a full quantization method with optimizing the transformation matrix in Winograd has been proposed \citep{tianqi2023towards}, which successfully restores model accuracy when employing the Winograd F($4\tiny{\times}4$, $3\tiny{\times}3$) algorithm with int8 quantization. Nevertheless, the methods above tend to struggle to achieve satisfactory accuracy recovery under sub-int8 quantization.

\textbf{Numerical Accuracy for Fastconv.} Another approaches focus on improving the intrinsic properties of the fast algorithm itself. As Winograd algorithms can be defined by root points, a bilinear approach that strikes a balance between computational complexity and numerical accuracy has been proposed \citep{barabasz2019winograd}. Additionally, two existing works \citep{barabasz2020error,alam2022winograd} aimed to discover more effective polynomial root points to improve numerical accuracy. The Winograd algorithms have also been extended to the Residue Number System (RNS) \citep{liu2020efficient}, decomposing single high-precision intermediate multiplications into multiple low-precision arithmetics (e.g., 8-bit). However, these all come at the cost of increased computational complexity.

\section{Preliminaries}
\textbf{Algorithms Construction.} Fast convolution algorithms, including Winograd, Fourier Transform, and Number Theoretic Transform all employ a three-stage computing process: transformations of filters and inputs, element-wise multiplication, and a transformation for generating outputs. 
The generalized form for fast 2D convolution is as follows:
\begin{equation}
    y=A^T[[GfG^T]\odot[B^TxB]]A
\end{equation}
where $\odot$ denotes element-wise multiplication, and $B$, $G$, and $A$ represent the linear transformations of the input, filter, and multiplication result. 

For one specific algorithm (whether Winograd, FFT or NTT), the $G$, $B$ and $A$ are all derive from a Vandermonde matrix $V$, which consist of a set of root points $s_{0}..s_{N-1}$: 
\begin{equation}
    V=
\begin{bmatrix}
 1 & s_0^1 & s_0^2 & ... &s_0^{N-1} \\
 1 & s_1^1 & s_1^2 & ... &s_1^{N-1}  \\
 .. & .. & .. & ... & .. \\
 1 & s_{N-1}^1 & s_{N-1}^2 & ... & s_{N-1}^{N-1}
\end{bmatrix}
\end{equation}
A $N\times N$ matrix $V$ and its inverse $V^{-1}$ can construct a fast convolution algorithm for $R\times R$ filter accommodating $N\times N$ inputs with $M\times M$ outputs, or $M\times M$ inputs with $N\times N$ outputs, where $N=M+R-1$.

\textbf{Difference.} The fundamental difference among various algorithms lies in the number field of $V$ and the chosen $S_n$. In the Winograd algorithm (also known as Toom-Cook algorithm), the $S_n$ are $N$ interpolation points selected in the real number field $\mathbb{R}$. Similarly, all arithmetic operations are performed in the $\mathbb{R}$. As a comparison, all arithmetic in Fourier convolution is defined in the complex field $\mathbb{C}$. And the matrix $V$ is the discrete Fourier transform matrix, where $S_n=e^{-j\frac{2\pi*n}{N}}$. The number theoretic transformation is similar in structure to the Fourier transform, but it operates in a finite field denoted as $\mathbb{F}_p$.

\textbf{Arithmetic Complexity Reduction.} Assuming these transformations are lightweight compared to element-wise multiply and can be amortized due to channel reuse, the fast algorithms consume $N^2=(M+R-1)^2$ multiplications to generate $M^2$ outputs, where the arithmetic complexity reduction is $\frac{M^2R^2}{(M+R-1)^2}$. However, convolution operations in CNNs are generally defined in $\mathbb{R}$, so employing fast algorithms defined in $\mathbb{C}$ or $\mathbb{F}_p$ (such as FFT and NTT) would lead to waste in the calculation. Hence, Winograd, defined in $\mathbb{R}$, is the most popular algorithm for model acceleration.

\textbf{Precision Requirement.} For Winograd, the extremum of a row in $V$ is $[1, S_n^{N-1}]$. So the required arithmetic precision grows exponentially with $N$. Thus only the Winograd algorithm with small $N$ is practical. In comparison, the FFT method performs a significant numerical advantage when dealing with large filters due to its numerically stable $V$ matrix. However, performing accurate Fourier transforms necessitates high-precision arithmetic. NTT method provide a bit-correct result for integral convolution. However, when using NTT to perform a calculation with $N$-bit inputs and $2N$-bit outputs, the transformed inputs must have a datawidth of at least $2N$-bit, which increases the datawidth of $\odot$. 

In summary, no matter which type of algorithm is chosen, achieving both robust computation and significant arithmetic reduction under low-precision operations remains a challenge.


\section{Symbolic Fourier Convolution Algorithm}

It is worth noting that the Fourier transform has better numerical stability, as all its root points are distributed on a circle of radius 1 in the complex plane. When dealing with larger $N$, it is more accurate than Winograd. However, Fourier transform has two serious drawbacks. First, its irrational coefficients are not friendly for low-precision format and would give more computation burden for transformation calculation.


In addition, the FFT is not as efficient as Winograd. There are two reasons for this. Firstly the FFT is computed using complex numbers and even after utilizing the Hamiltonian symmetry with real sequences and the fast complex multiplication, each complex multiplication still requires 1.5 real multiplications. Secondly, the direct calculation result of the FFT is a circular convolution, so padding the sequence with zeros to achieve a linear convolution is required, which would wasted computation. 


We propose two key improvement strategies to address these drawbacks:
\begin{itemize}
    \item [1.] We employ symbolic computation rather than numerical computation to implement the discrete Fourier transform (DFT). By selecting an appropriate number of DFT points, we can avoid or minimize the irrational values involved in computing. All complex points can be represented by first order integer coefficient polynomials under both 4 and 6 DFT points.
    \item [2.] We introduce correction terms to fully exploit the cyclic convolution output generated by the Fourier method to enhance computing utilization, and the smaller number of transformation points we chose also helps to reduce the proportion of complex arithmetic.
\end{itemize}

\subsection{Symbolic Computing for DFT}

Generally, the coefficients of the N-point DFT are derived from: $$e^{\frac{2\pi n}{N}j}=cos(\frac{2\pi n}{N})+jsin(\frac{2\pi n}{N}), n=0,1,..,N-1$$

when $\frac{n}{N}\notin\{0,\frac{1}{4},\frac{1}{2},\frac{3}{4}\}$, irrational values will be introduced. To eliminate the rounding errors that arise from these irrational values, we employ symbolic computation rather than numerical calculation. This approach represents irrational values in polynomials with integer coefficients.

To illustrate, we consider the 3-point DFT. For a real input sequence $x=(x_0,x_1,x_2)^T$, its transform processing can be represented as:
\begin{equation}
\label{eq:DFT3}
\begin{bmatrix}
 X_0\\
 X_1\\
 X_2\\
\end{bmatrix}=
\begin{bmatrix}
1&1&1\\
 1&s&s^2\\
 1&s^2&s\\
\end{bmatrix}\begin{bmatrix}
 x_0\\
 x_1\\
 x_2\\
\end{bmatrix} , s=e^{\frac{2\pi}{3}j}
\end{equation}

\begin{figure} [h]
	\begin{center}
	\includegraphics[width=0.5\textwidth]{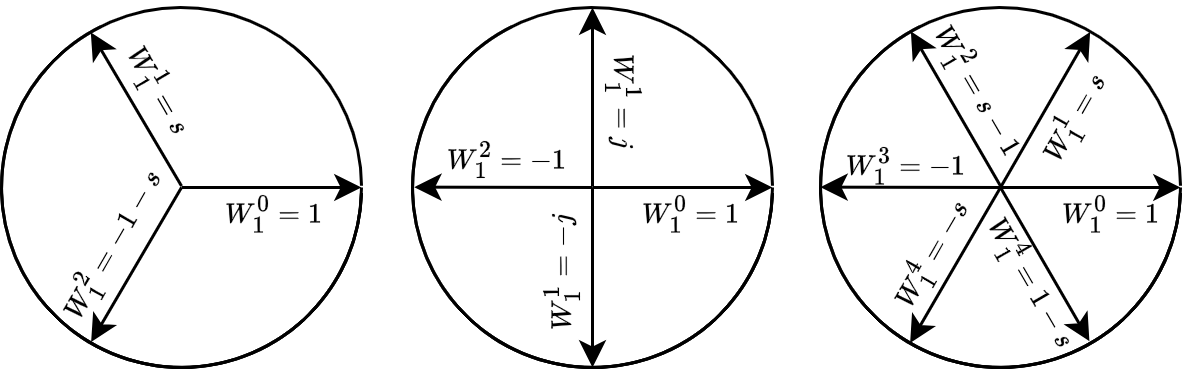}
	\caption{The geometric symmetry in DFT-3, DFT-4 and DFT-6.} 
	\label{fig::symmetry}
        \end{center}
\end{figure} 

We do not substitute the numerical value of $s$ into the calculation. Instead, we represent and compute subsequent variables using the polynomial form of s. This allows us to express the DFT outputs $X_n$ as $X_n=X_{n,0}+X_{n,1}\cdot s+X_{n,2}\cdot s^2$.

By exploiting the geometric symmetry among $1,s,s^2$, the $s^2$ can be expressed as $s^2=-(s^0+s^1)$, as Figure \ref{fig::symmetry}. Further, the Hamiltonian symmetry of the real signal Fourier transform can reduce the number of components by almost half. Thus we can rearrange Equation \ref{eq:DFT3} as following:

\begin{equation}
\begin{bmatrix}
X_0\\
X_1\\
X_2\\
\end{bmatrix}=\begin{bmatrix}
 1 & 0 & 0\\
  0 & 1 & s\\
  0 & 1 & s^2\\
\end{bmatrix}
\begin{bmatrix}
1&1&1\\
 1&0&-1\\
 0&1&-1\\
\end{bmatrix}\begin{bmatrix}
 x_0\\
 x_1\\
 x_2\\
\end{bmatrix}
\end{equation}

In the above formula, $ s=e^{\frac{2\pi}{3}j}$ is not need to be explicitly included in the calculation but serves as a notation from the outset. Similar to complex number, the production of $a_0+a_1s$ and $b_0+b_1s$ can be calculated as:




\begin{equation}
\label{eq:DFT3M}
\begin{aligned}
    (a_0+a_1s)&(b_0+b_1s)=c_0+c_1s, \quad\begin{bmatrix}
 c_0 \\
 c_1
\end{bmatrix}= \\ \begin{bmatrix}
 1 & -1 &0\\
 1& 0 &1
\end{bmatrix}&(\begin{bmatrix}
 1 & 0\\
 0& 1 \\
 -1 & 1
\end{bmatrix}\begin{bmatrix}
 a_0\\
 a_1
\end{bmatrix}\odot \begin{bmatrix}
 1 & 0\\
 0& 1 \\
 1 & -1
\end{bmatrix}\begin{bmatrix}
 b_0\\
 b_1
\end{bmatrix})
\end{aligned}
\end{equation}



Through enumeration, we can identify that $6$ and $4$ are suitable choices for DFT points in small-size convolution applications. 

\emph { 1) DFT-6}

Considering DFT-6, its transformation coefficients consist of six values: $1, e^{j\frac{\pi}{3}}, e^{j\frac{2\pi}{3}}, -1, e^{j\frac{4\pi}{3}}, e^{j\frac{5\pi}{3}}$. Defining $s = e^{\frac{\pi}{3}j}$, thus $s^2=s-1$, which allows all coefficients to be expressed as first-order polynomials of $s$: $1,s,s-1,-1,-s,1-s$. When two first-degree polynomials are multiplied, any quadratic term can be reduced to a first-degree term using the rule $s^2=s-1$. 
Therefore, the DFT-6 transform processing under symbolic computation is as follows:
\begin{equation}
\label{eq:DFT6}
\begin{aligned}
   DFT6(x)=S_6F_6x=
   \begin{bmatrix}
 1 & 0 & 0 & 0 & 0 & 0\\
 0 & 1&  s& 0& 0& 0\\
 0 & 0 & 0 & 1 & s& 0\\
 0& 0 & 0&  0 & 0& 1\\
 0 & 1& -s^2& 0& 0& 0\\
 0 & 0 & 0 & 1& -s^2 & 0 \\
\end{bmatrix} \cdot 
\\ \begin{bmatrix}
 1 & 1 & 1 & 1 & 1 & 1\\     
 1 & 1 & 0 & -1& -1& 0\\    
 0 & -1& -1& 0& 1 & 1\\    
 1 & 0 & -1 &  1& 0 &-1\\    
 0 & -1& 1 &  0& -1& 1\\
 1& -1 & 1& -1 & 1& -1\\     
\end{bmatrix} \cdot
\begin{bmatrix}
 x_0\\
 x_1\\
 x_2\\
 x_3\\
 x_4\\
 x_5\\
\end{bmatrix}
\end{aligned}
\end{equation}

Here, $S_6$ represents the format transition from symbolic to numerical computing without any arithmetic, and $F_6$ signifies the Fourier transform under symbolic computing. We refer to the intermediate matrix as SFT-6 (Symbolic Fourier Transform-6), as its coefficients consist solely of 1, -1, and 0. The inverse transform iDFT6 can be rearranged in a similar way. $F_6$ has its fast algorithm by decomposing it into DFT-3 and DFT-2. Therefore, only 14 addition operations are needed to perform SFT-6 (subtraction can be considered as addition of complement). The inverse transformation matrix $iF_6$ can be derived in a similar way: 
\begin{equation}
iF_6=\frac{1}{6}
\begin{bmatrix}
 1 & 1 & 1 &1 &1 & 1  \\
 1 & -1 & -2 & -1 & 1 & 2\\
 -1 & -2 & -1 &1 & 2 & 1\\
 -1 & -1 & 2 &-1 & -1 & 2\\
 -2 & 1 & 1 &-2 & 1 & 1\\
  -1 & 1 & -1 &1 & -1 & 1\\
\end{bmatrix}
\end{equation}
Note that $\frac{1}{6}$ can be fused into floating model without having to compute that division operation during inference stage.

In the element-wise multiplication steps, multiplications are performed in polynomial form. Multiplying two 1st-order polynomials requires 4 real number multiplications. We can utilize a short fast convolution algorithm to reduce it to 3:

\begin{equation}
\label{eq:DFT6M}
\begin{aligned}
    (a_0+a_1s)&(w_0+w_1s)=o_0+o_1s, \quad\begin{bmatrix}
 o_0 \\
 o_1
\end{bmatrix}=\\ \begin{bmatrix}
 1 & -1 &0\\
 -1& 0 &1
\end{bmatrix}&(\begin{bmatrix}
 1 & 0\\
 0& 1 \\
 1 & 1
\end{bmatrix}\begin{bmatrix}
 a_0\\
 a_1
\end{bmatrix}\odot \begin{bmatrix}
 1 & 0\\
 0& 1 \\
 1 & 1
\end{bmatrix}\begin{bmatrix}
 w_0\\
 w_1
\end{bmatrix})
\end{aligned}
\end{equation}

\emph { 2) DFT-4}

Similarly, the DFT-4 can be constructed in the same manner.
\begin{equation}
\label{eq:DFT4}
\begin{aligned}
 DFT4(x)=S_4F_4x=\begin{bmatrix}
 1 & 0 & 0 & 0 \\
 0 & 1& j&  0 & \\
  0& 0 & 0& 1 \\
  0& 1 & -j& 0 \\
\end{bmatrix} \cdot \\
\begin{bmatrix}
 1 & 1 & 1 & 1 \\
 1 & 0 & -1 & 0\\
 0 & -1 & 0&  1\\
 1& -1 & 1& -1 \\
\end{bmatrix}  \begin{bmatrix}
 x_0\\
 x_1\\
 x_2\\
 x_3\\
\end{bmatrix}
\end{aligned}
\end{equation}

The multiplication on the DFT-4 convolution can be performed as:
\begin{equation}
\label{eq:DFT4M}
\begin{aligned}
(a_0+a_1s)&(w_0+w_1s)=o_0+o_1s, \quad\begin{bmatrix}
 o_0 \\
 o_1
\end{bmatrix}=\\ \begin{bmatrix}
 1 & -1 &0\\
 -1& -1 &1
\end{bmatrix}&(\begin{bmatrix}
 1 & 0\\
 0& 1 \\
 1 & 1
\end{bmatrix}\begin{bmatrix}
 a_0\\
 a_1
\end{bmatrix}\odot \begin{bmatrix}
 1 & 0\\
 0& 1 \\
 1 & 1
\end{bmatrix}\begin{bmatrix}
 w_0\\
 w_1
\end{bmatrix})
\end{aligned}
\end{equation}

If we want to compute $A((Gf)\odot (Bx))$ directly in the real number field, akin to the Winograd algorithm, without involving polynomial multiplication, we can integrate Eq. \ref{eq:DFT6M} or Eq. \ref{eq:DFT4M} into the SFT matrix Eq. \ref{eq:DFT6} or Eq. \ref{eq:DFT4}. In the 1D case, this does not impact efficiency. However, in the 2D case, it introduces slight redundant components and marginally reduces the acceleration ratio. We list these algorithms with polynomial multiplication integrated in the appendix.

\subsection{Achieving Efficient Linear Convolution}
The conventional Fourier Transform method inherently generates cyclic convolution. As a consequence, only $(N-R+1)^2$ components are valid for the intended $N^2$ linear convolution. However, it's noteworthy that the invalid results are not entirely useless. They actually contain partial sums that can be effectively utilized. By intelligently applying correction terms to these partial sums, it is possible to convert them into valid outputs. This approach significantly enhances the efficiency of the convolution process.

\begin{figure*} [h]
	\begin{center}
	\includegraphics[width=0.8\textwidth]{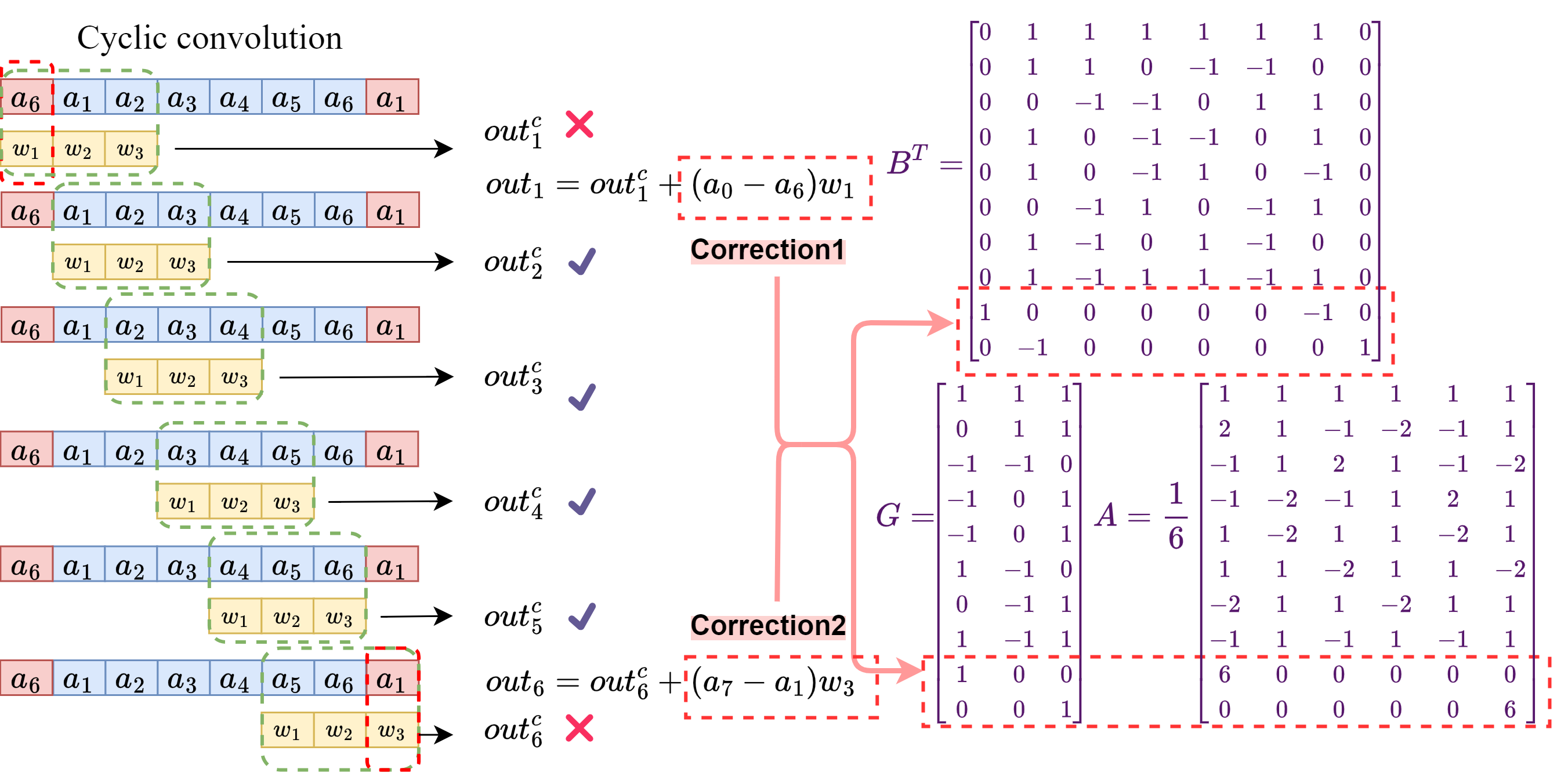}
	\caption{Converting invalid cyclic convolution to linear convolution, which is a process that can also be used to adjust the input tile sizes.} 
	\label{fig::convmodify}
        \end{center}
\end{figure*}

Figure \ref{fig::convmodify} illustrates an example of Fourier-based cyclic convolution for $N=6$ and $R=3$. The first term $o_1^c$ is equal to $a_6w_1+a_1w_2+a_2w_1$, but the desired output is $o_1=a_0w_1+a_1w_2+a_2w_1$. To align $o_1$ with $o_1^c$, a corrective term is introduced to obtain the desired output: $o_1 = o_1^c + (a_0 - a_6)w_1$. This adjustment allows us to obtain an additional correct result by adding just one MAC operation, thus utilizing the Fourier convolution output more efficiently compared to discarding erroneous terms. 

To unambiguously represent a particular algorithm, we employ the notation SFC-$N$($M$, $R$), where $N$ represents the length of the SFT transformation, $M$ represents the feature tile size, and $R$ represents the kernel size. For example, the SFC-6(6×6, 3×3) algorithm is constructed based on a $6$-point Fourier transform, employing a 3×3 kernel size and a 6×6 feature tile size.

By introducing correction terms, we can also adjust input tile size $M$ independently. For example, as the images in the ImageNet dataset are in size 224×224, the feature map size in the model has a common factor of 7.  
Utilizing the SFC-6(7×7, 3×3) algorithm to infer models trained on Imagenet would have higher tiling efficiency without the need for zero padding. The transformation matrix integrated polynomial multiplication of the SFC-6(7×7, 3×3) is as follows:
\begin{equation}
\begin{aligned} {\tiny
B^T=\begin{bmatrix}
0 & 1 & 1 & 1 & 1 & 1 & 1&0&0\\     
0 & 1 & 1 & 0 & -1& -1& 0&0&0\\    
0 & 0 & -1& -1& 0& 1 & 1&0&0\\    
0 & 1 & 0 & -1& -1& 0 & 1&0&0\\   
0 & 1 & 0 & -1 &  1& 0 &-1&0&0\\    
0 & 0 & -1& 1 &  0& -1& 1&0&0\\
0 & 1 & -1& 0 &  1& -1& 0&0&0\\
0 & 1& -1 & 1& -1 & 1& -1&0&0\\   
1 & 0  & 0 & 0  & 0&  0& -1&0&0\\
0 & -1 & 0 & 0  & 0&  0& 0 &1&0\\
0 & -1 & 0 & 0  & 0&  0& 0 &1&0\\
0 & 0  & -1& 0  & 0&  0& 0 &0&1\\

\end{bmatrix} },\\
{\tiny A = \frac{1}{6}\begin{bmatrix}
 1 & 1 & 1 &1 &1 & 1 & 1 \\
  2 & 1 & -1 &-2 &-1 & 1 & 2 \\
 -1 & 1 & 1 & 1 & -1 & -2& -1\\
 -1 & -2 & -1 &1 & 2 & 1& -1\\
 1 & -2 & 1 &1 & -2 & 1& 1\\
 1 & 1 & -2 &1 & 1 & -2& 1\\
 -2 & 1 & 2 &-2 & 1 & 1& -2\\
  -1 & 1 & -1 &1 & -1 & 1& -1\\ 
  6& 0  &0  &  0 &0 & 0  & 0\\
  0& 0  &0  &  0 &0 & 6  & 0\\
  0& 0  &0  &  0 &0 & 0  & 6\\
  0& 0  &0  &  0 &0 & 0  & 6\\
\end{bmatrix}} ,\\
{\tiny
G=\begin{bmatrix}
 1 & 1 & 1 \\
 0 & 1 & 1\\
-1 & -1& 0\\
-1& 0& 1\\
-1& 0& 1\\
1 & -1 & 0\\
0 & -1& 1 \\
1 & -1&  1\\
1&0&0\\
0&  0 & 1 \\
 0 &1&0\\
 0 & 0 &1 \\
\end{bmatrix}}
\end{aligned}
\end{equation}

The SFC-6(6×6, 3×3) algorithm can reduce 73\% multiplications for 3×3 convolutions. And multiplications can be optimized by 81\% and 79\% for 5×5 and 7×7 convolutions, respectively. A selection of achievable Symbolic Fourier Convolution algorithms is listed in Table \ref{tab::algscompare}. For comparison, we also list their efficiencies and the numerical errors obtained from simulation and theoretical analyses (Detailed in Section \ref{error_analysis}). In Table \ref{tab::algscompare}, we can see that for 3×3 convolution the SFC-$6$($6\tiny{\times}6$, $3\tiny{\times}3$) algorithm is 1.64× faster than the Winograd (2×2, 3×3) algorithm, while maintaining nearly the same numerical error. 

Further, for larger kernel size, the Winograd algorithm is no longer able to construct linear transformations with sufficiently low numerical errors, no matter how the root points are chosen. Although there exists work by splitting the core, making it possible to compute large size convolutions with the Winograd(2×2, 3×3) algorithm with low numerical error \citep{huang2020dwm}, it is not possible to achieve a lower arithmetic complexity than Winograd(2×2, 3×3). Whereas our algorithm provides higher speedups and keeps the high numerical stability.

\begin{table}[h]
    \small
    \begin{center}
     \caption{Comparison of different fast convolution algorithms. Mean square error was measured on randomly generated data, and $\kappa(A^T)$ was calculated using the singular value of $A^T$.} 
    \begin{tabular}{c|c|c|c}
\hline
        Algorithm & \begin{tabular}[c]{@{}c@{}}\ Mean Square\\ Error\end{tabular} &$\kappa(A^T)$ &\begin{tabular}[c]{@{}c@{}}\ Arithmetic \\ Complexity\end{tabular}  \\ \hline 
        direct convolution & 1.0 &1.0 & 100\% \\ \hline
        Wino($2\tiny{\times}2$, $3\tiny{\times}3$) & 2.2 &2.4& 44.4\% \\ \hline
        Wino($3\tiny{\times}3$, $3\tiny{\times}3$) & 6.4 &14.5& 30.4\% \\ \hline
        Wino($4\tiny{\times}4$, $3\tiny{\times}3$)  & 10.5 &20.1& 25\% \\ \hline
        \rowcolor{yellow!30} SFC-4($4\tiny{\times}4$, $3\tiny{\times}3$) & 2.4  &2.7& ~31.94\% \\ \hline
        \rowcolor{yellow!40} SFC-6($6\tiny{\times}6$, $3\tiny{\times}3$) & 2.4 &3.3& 27.16\%\\ \hline
        \rowcolor{yellow!50} SFC-6($7\tiny{\times}7$, $3\tiny{\times}3$)  & 2.6 &3.4& 29.93\% \\ \hline \hline
        Wino($2\tiny{\times}2$, $5\tiny{\times}5$)  & 10.5 & 20.1&36\% \\ \hline
        \rowcolor{yellow!40} SFC-6($6\tiny{\times}6$, $5\tiny{\times}5$)  & 3.6 &3.5& 20.44\%\\ \hline \hline
        Wino($2\tiny{\times}2$, $7\tiny{\times}7$) & 28.1 &31.0 &32.6\% \\ \hline
        \rowcolor{yellow!40} SFC-6($4\tiny{\times}4$, $7\tiny{\times}7$)  & 3.6 &3.5 &21.99\% \\ \hline
    \end{tabular}
   
    \label{tab::algscompare}
    \end{center}
\end{table}


\section{Error Analysis and Frequency-wise Quantization} \label{error_analysis}


This section would analyze the numerical error of fast convolution algorithms, which can be used to guide the development of quantization methods and serve as a benchmark for assessing the numerical stability across various fast algorithms.

To cover the direct convolution into the same error analysis model, we can consider it as a fast convolution with $R=3$, $M=1$.
For derivation convenience, we use the overlapped output form, in which $A$ is a reversible square matrix.
\begin{equation}
    \label{eq::dc}
   \begin{bmatrix}
 y_0 \\
 y_1 \\
 y_2
\end{bmatrix}^T=
\begin{bmatrix}
 1 &  & \\
  & 1 & \\
  &  & 1
\end{bmatrix}\cdot((
\begin{bmatrix}
 1 &  & \\
  & 1 & \\
  &  & 1
\end{bmatrix}\cdot 
\begin{bmatrix}
 f_0 \\
 f_1 \\
 f_2
\end{bmatrix}
)\odot (\begin{bmatrix}
 1
\end{bmatrix}\cdot \begin{bmatrix}
 x_0
\end{bmatrix}))
\end{equation}

We denote the quantized element-wise multiply as $\odot_Q$, through which the operands are quantized and multiplied. $\delta y$ represents the calculation error of the output $y$. The error forward propagation process can be described as:
\begin{equation}
    \label{eq::qfca1d}
    y+\delta y=A^T\cdot((G\cdot f)\odot_Q (B^T\cdot x))
\end{equation}
We assume these transformations to be accurate, and the quantized operator $\odot$ introduces rounding errors. These errors would be subsequently magnified by the matrix multiplying $A^T\cdot$. 

Set $s+\delta s=(G\cdot f)\odot_Q (B^T\cdot x)$, by substituting $y=A^T\cdot s$ into the Equation \ref{eq::qfca1d}, we can obtain:
\begin{equation}
    \label{eq::qfcax}
     \delta y=(A^T)^{-1}\cdot \delta s
\end{equation}

Applying the properties of the paradigm $||*||$ yields:
\begin{equation}
    \label{eq::xB2}
        ||\delta y||<=||(A^T)||\cdot||\delta s||
\end{equation}
Similar, bringing in $||s||<=||(A^T)^{-1}||\cdot||y||$, we have: 
\begin{equation}
    \label{eq::xB3}
        \frac{||\delta y||}{||y||}<=||(A^T)^{-1}||\cdot||(A^T)||\cdot \frac{||\delta s||}{||s||}
\end{equation}
Here we can perform an analysis of Eq.\ref{eq::xB3}. 
The first term $||(A^T)^{-1}||\cdot||(A^T)||$ is the condition number of matrix $A^T$, denoted as $\kappa(A^T)$. This term indicates the amplification factor applied to the error $\delta s$.
The condition number of the standard orthogonal matrix is constant 1, like $A$ in direct convolution (Eq.1) and vanilla Fourier convolution, which would not amplify $\delta s$ anymore.
While the $\kappa(A^T)$ for Winograd convolution can reach up to 31.0, as listed in Table \ref{tab::algscompare}. 
However, in our method, the $\kappa(A^T)$ takes the values of 2.7, 3.3, and 3.5 for SFC4(4,3), SFC6(6,3), and SFC6(4,7), respectively, which is much less than that of Winograd. 

The final term $\frac{||\delta s||}{||s||}$ in Eq.\ref{eq::xB3} is the error caused by the quantized operator $\odot_Q$, which depends on the data width, quantization method, and data distribution.
For floating numbers, the error is fairly stationary as every operand has its own exponent code(scaling factor). We conduct numerical experiments under single precision (fp16) in Table \ref{tab::algscompare}, and it is find that the numerical error is highly correlated to the $\kappa(A^T)$, in accordance with our analysis.

\begin{figure} [h]
	\begin{center}
	\includegraphics[width=0.5\textwidth]{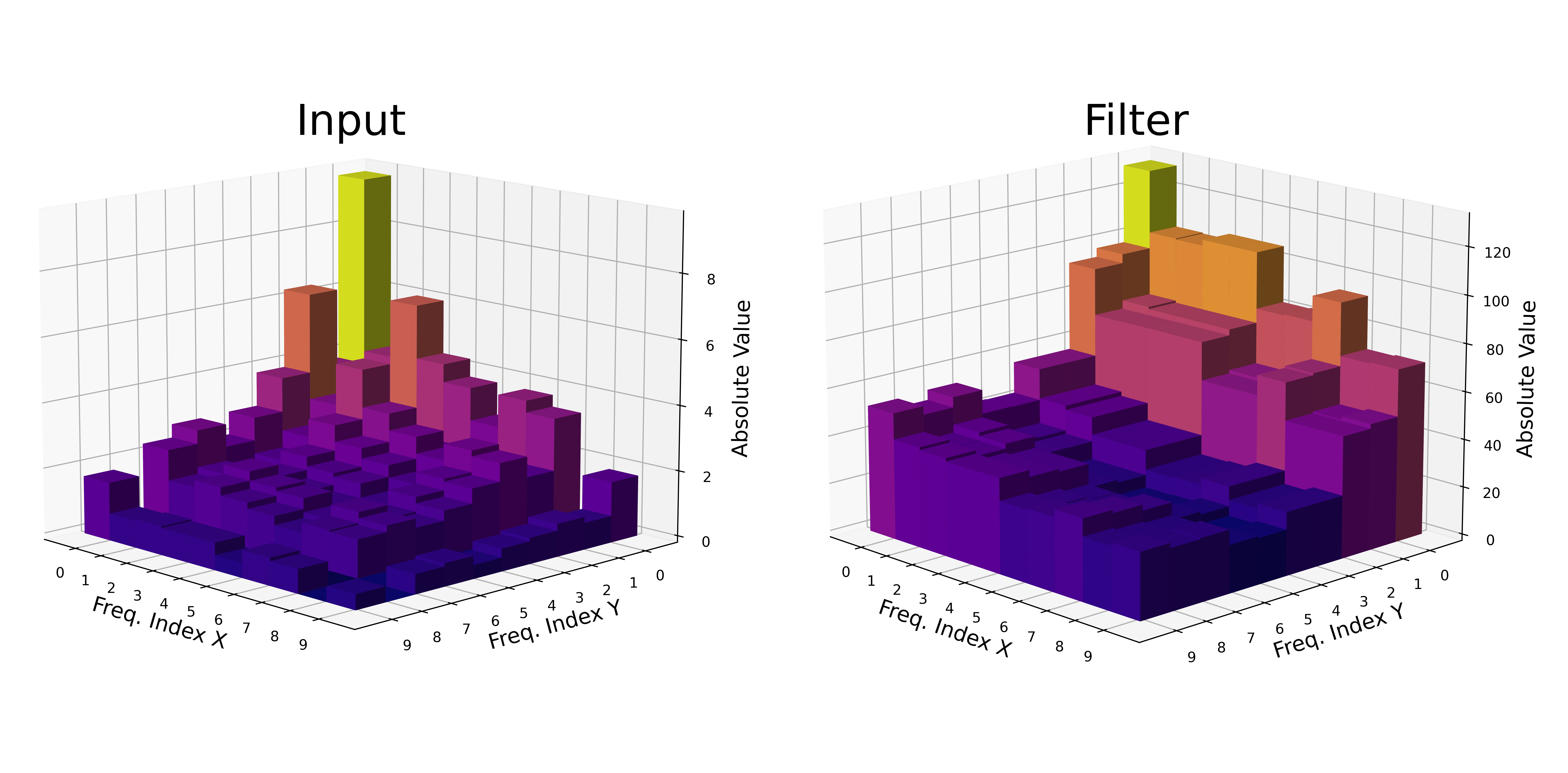}
	\caption{The energy distribution in the frequency domain of the 9-th layer in Resnet18. Energy is concentrated in the low frequencies.} 
	\label{fig::distrub}
        \end{center}
\end{figure} 

For model quantization, numbers in a group share the same scaling factor to achieve integer representation. The granularity of the scaling factor group should fit with the data distribution to achieve low quantization error.
Some work has found that in Winograd convolution, scaling factor grouping based on the transformation domain coordinates can recover model accuracy effectively. Here we give a brief theoretical explanation. Assume that the input $x$ has a fixed energy $||x||=1$, the maximum possible value in the transformation domain is $(1,1,1,8,8,1)$ for F(4,3) algorithm, which would be quadratic in 2D convolution. Grouping one tensor to one scaling factor would cause a waste of 3/6 bit in 1D/2D. For Fourier convolution, each frequency would have the same maximum value. However, due to the frequency properties of the model inputs (image or audio), there will be a tendency for the energy to converge towards lower frequencies, as Figure \ref{fig::distrub} shows. Thus grouping scaling factors by frequency can also reduce the error of Fourier convolution, but not as necessary as Winograd. This is because the aforementioned tendency is only significant in the first few layers, and the magnitude variance is not as large as Winograd.
\begin{equation}
\small
   y=\sum_{C_{in}}(s_{Tx}\left \lceil  B^TxB/s_{Tx} \right \rfloor_{intN_{Tx}}\odot s_{Tf}\left \lceil GfG^T/s_{Tf} \right \rfloor_{intN_{Tf}}) 
\end{equation}

The scaling factor $s_{Tx}$ is of size $[T\tiny{\times} T]$, where $T$ is the size of the transform domain. For the scaling factor $s_{Tf}$ of weights, considering that per-channel quantization can achieve better results in direct convolutions, we suggest combining per-frequency quantization and per-channel quantization whose $s_{Tf}$ is of size $[OC\tiny{\times}T\tiny{\times} T]$ to achieve higher accuracy.

\section{Experimental Evaluation }
We conducted experiments on image classification tasks to demonstrate the effectiveness of our algorithms. 
To comprehensively evaluate the computation cost of models accelerated by fast convolution algorithm and quantization, it is critical to consider both the reduction in arithmetic complexity afforded by fast convolution and the reduction in arithmetic data width introduced by quantization. Consequently, we adopt bit-operations (BOPs) as a metric of computation cost, diverging from the traditional floating-operations (FLOPs). 
In this metric, an n-bit addition operation requires n BOPs, whereas an n-bit multiplication costs n(n-1) BOPs. This is because an n-bit multiplication can be decomposed into n-1 instances of n-bit additions. The transformation cost of fast algorithms is also taken into account.  


\subsection{Post-training Quantization on Image Classification Benchmarks}

Experiments were conducted on the ImageNet dataset, which contains 1.4 million images of size 224×224×3, distributed across 1,000 classes. We randomly selected 500 images from training set to create the calibration set for PTQ fine-tuning. Model accuracy was evaluated on the validation set. We utilized pre-trained fp32 models from TorchVision as our benchmarks. All batch normalization layers were fused into the preceding convolution layers prior to quantization.

We conducted quantization on the following algorithms: 1) Direct convolution, 2) The well-known Winograd F(4×4, 3×3) algorithms, which have been extensively researched for their quantization methods in recent years, and 3) our proposed SFC algorithms, including 1D and 2D format. For all these methods, all 3×3 convolution layers with a stride of 1 were replaced by the corresponding algorithm. Direct convolution and our SFC algorithms were quantized using AdaQuant \citep{Hubara_2020}, while the Winograd algorithm was processed with Scaling Gradient Backward \citep{jain2020trained}, due to observed convergence differences with AdaQuant in the Winograd F(4×4, 3×3). All the data in the spatial domain are quantized to int8, and the data in the transformation domain are quantized from int8 to int4. As mentioned in Section \ref{error_analysis}, storing spatial domain data in external storage is unnecessary; instead, data can be stored in the transform domain to avoid errors caused by multiple quantization operators.




\begin{figure} [ht]
\centering
\includegraphics[width=0.49\textwidth]{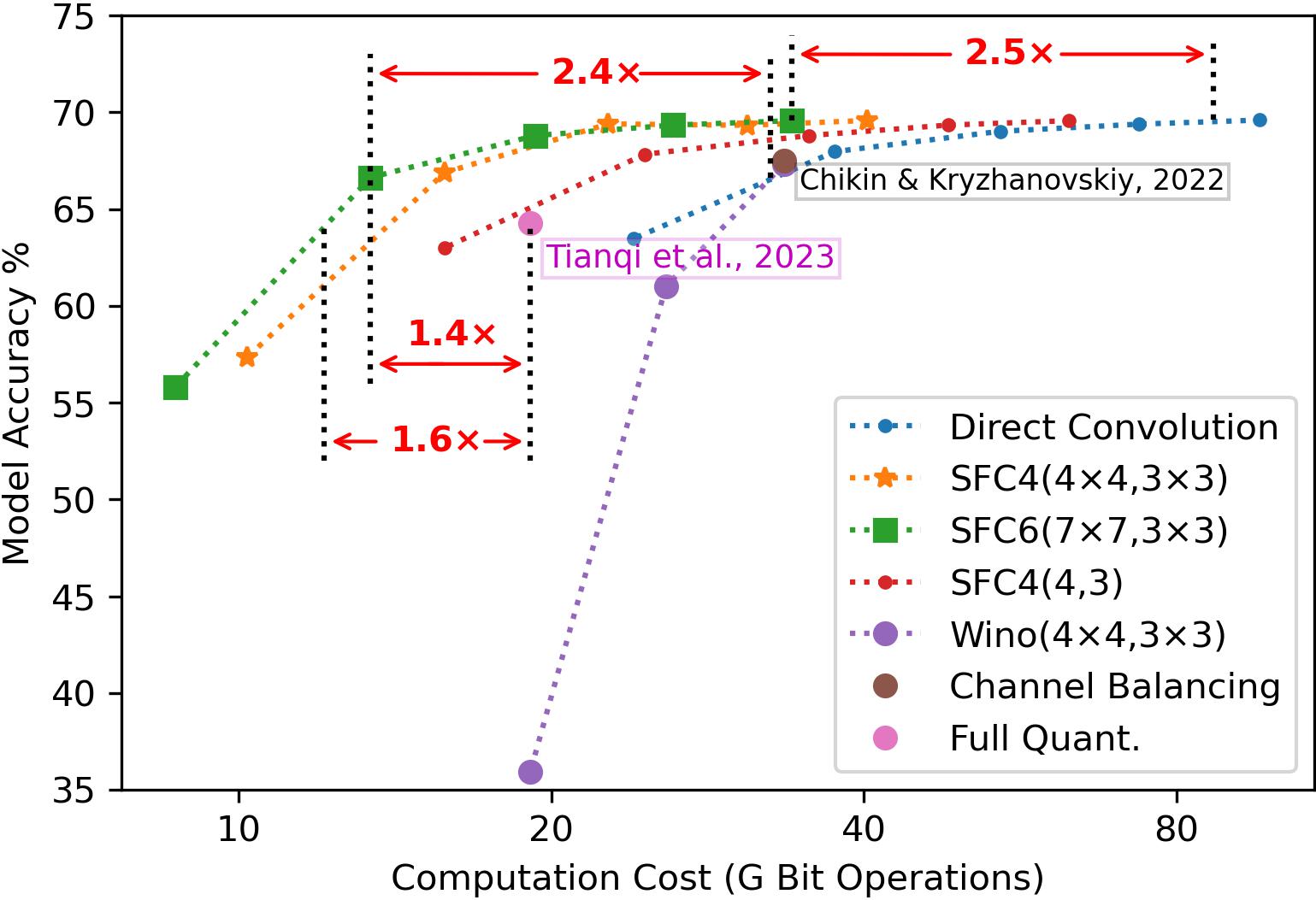}
\caption{Model Accuracy vs. Computation Cost. We done post-training quantization for Resnet18 on Imagenet dataset from int8 to int4.}
\label{fig::curve}
\end{figure}

\begin{table} [h]
  \small
  \caption{Post-training quantization experiments on ImageNet.}
  \label{table:experiment}
  \begin{center}
  \begin{tabular}{ c c c c c  c c }
    \hline
   
    Resnet18 & &Bits &Top-1 &$\Delta$\\
    \hline
    Channel Balancing&Wino(4×4, 3×3) &8&67.54 & -2.22\\
    Full Quant.&Wino(4×4, 3×3)&8&68.16& -1.60\\
    Full Quant.&Wino(4×4, 3×3) &6&64.34& -5.42\\
    \rowcolor{yellow!40} Ours&SFC6(7×7, 3×3)&8&\bf{69.59}& \bf{-0.17}\\
    \rowcolor{yellow!70} Ours&SFC6(7×7, 3×3)&6&\bf{68.80} & \bf{-0.96}\\
    \hline
    Resnet34 & &Bits &Top-1 &$\Delta$\\
    \hline
    Channel Balancing&Wino(4×4, 3×3)&8&71.86& -1.44\\
    Full Quant.&Wino(4×4, 3×3)&8&71.75 & -1.55\\
    Full Quant.&Wino(4×4, 3×3)&6&68.80 & -4.50\\
    \rowcolor{yellow!40} Ours&SFC6(7×7, 3×3) &8&\bf{73.14}& \bf{-0.16}\\
    \rowcolor{yellow!70} Ours&SFC6(7×7, 3×3) &6&\bf{72.40}& \bf{-0.90}\\
    
    \hline
    Resnet50  & &Bits &Top-1 &$\Delta$\\
    \hline
    Channel Balancing&Wino(4×4, 3×3)&8&75.84& -0.29\\
    Full Quant.&Wino(4×4, 3×3)&8&75.74& -0.40\\
    Full Quant.&Wino(4×4, 3×3)&6&74.75& -1.39\\
    \rowcolor{yellow!40}Ours&SFC6(7×7, 3×3)&8&\bf{76.02} &\bf{-0.12}\\
    \rowcolor{yellow!70}Ours&SFC6(7×7, 3×3) &6&\bf{75.54} & \bf{-0.60}\\
    \hline
  \end{tabular}
  \end{center}
\end{table}
Since it is not feasible to align a specific category of data, such as ensuring uniform computational costs across a column to compare accuracy in the table, we have illustrated the accuracy curves in relation to computation cost (BOPs) in Figure \ref{fig::curve} for a visual comparison of different algorithms. To ensure a fair and comprehensive comparison, we have included the state-of-the-art Winograd-quantization work in the figure, including Channel Balancing \citep{chikin2022channel} and Full Quantization \citep{tianqi2023towards}. A subset of the detailed data is presented in Table \ref{table:experiment}.  As depicted in Figure \ref{fig::curve}, the SFC-6(7×7, 3×3) algorithm demonstrates a computation cost reduction of \textbf{1.6× to 2.5×} compared to both the Winograd F(4×4, 3×3) and direct convolution algorithms, while maintaining equivalent model accuracy. The experiments confirm that the SFC algorithms achieve a reduction in arithmetic complexity comparable to Winograd, with model accuracy similar to that of direct convolution (quantization-alone).

\begin{figure} [h]
	\begin{center}
	\includegraphics[width=0.39\textwidth]{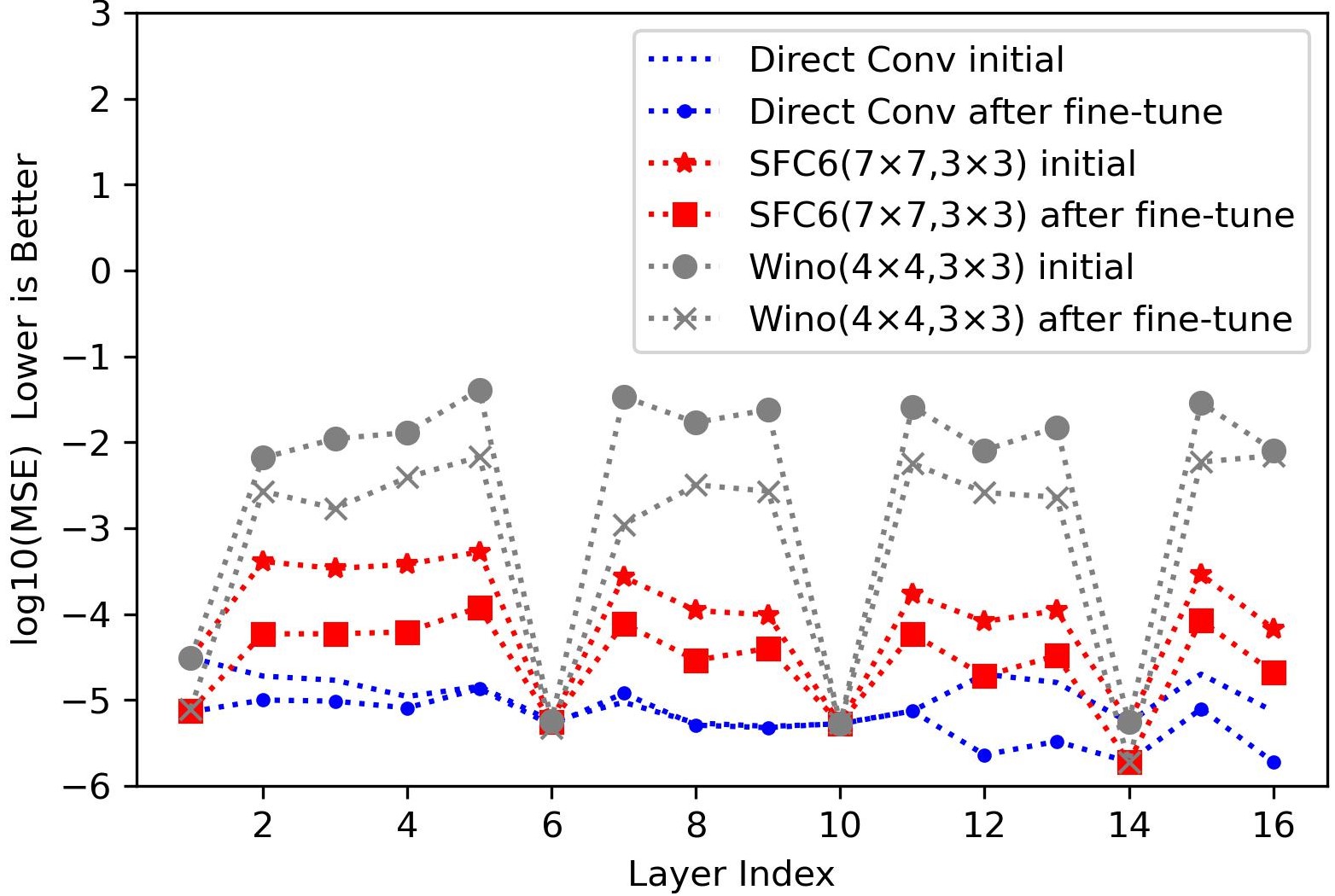}
	\caption{The MSE for different algorithms under int8 PTQ. Lower numerical error results to better PTQ results.} 
	\label{fig::mse}
        \end{center}
\end{figure} 

We also plot the mean squared error (MSE) between accelerated layers to fp32 layers for different methods under int8 quantization. The experimental results, presented in Figure \ref{fig::mse}, match perfectly with the numerical error analysis conducted in Section 5 and Table \ref{tab::algscompare}.

\subsection{FPGA Simulation}
We develop RTL code for the convolution accelerator based on the SFC-6(7×7,3×3) algorithm. The resource consumption and timing report are synthesized using Xilinx Vivado tools. The parallelism of our design is configured at [4×4×7×7], indicating that one convolution operator with 4 input channels, 4 output channels, and 7×7 feature map are processed simultaneously. The VGG-16 model is taken as an example, whose convolution layers all have 3×3 filters with stride=1, making it well-suited for fast convolution. All components in our datapath are quantized to int8, and all computing stages in fast convolution are designed to operate in a full pipeline architecture. The DSP48 hardcore is a crucial resource on FPGA, as it can efficiently deploy multiply operations. One DSP48 unit can implement two int8 multipliers or one int16 multipliers, which means that our implementation consumes only 1056 DSPs (calculated as 4×4×132×0.5). In comparison, Winograd-based or NTT-based accelerators \citep{Liang_2020,Prasetiyo_2023} require more DSPs for high precision multipliers, and accelerators for direct convolution \citep{9354493} need more DSPs due to a higher complexity of multiplications. The result highlights the efficiency of integrating fast convolution with model quantization.

\begin{table}[h]
\footnotesize
  \caption{FPGA synthesis results.}
  \label{table:fpga_experiment}
  \begin{center}
  \small
  \begin{tabular}{c c c c c c}
    \hline
    Works&\thead{\scriptsize{Liang et al.,}\\\scriptsize{2020}}&\thead{\scriptsize{Prasetiyo et al.,}\\\scriptsize{2023}}&\thead{\scriptsize{Huang et al.,}\\\scriptsize{2022}}& Ours\\
    \hline
    Algorithm&Winograd&NTT&direct conv&SFC\\
   
    Platform&zcu102&xc7vx980t&alveo U50&xczu19eg\\
   
   Precision&16bit&8bit/21bit&8bit&8bit\\
   LUTs&-&601.7K&335K&221K\\
   DSPs&2304&4100&3395&1056\\
   Clock&200M&200M&200M&200M\\
   \thead{\scriptsize{Throughput}\\ \scriptsize{(GOPs)}}&2601&2859.5&1000&2129\\
   \thead{\scriptsize{GOPs/DSPs}\\\scriptsize{/Clock}}&5.64&3.48&1.96& 10.08\\
    \hline
  \end{tabular}
  \end{center}
\end{table}

\subsection{Ablation Study on Quantization Granularity}
In Section \ref{error_analysis}, we theoretically predict that frequency-wise quantization will produce less error than tensor-wise quantization. Here, we provide an ablation experiment on Resnet-18 by enumerating combinations of different quantization granularity in Table \ref{table:ab_experiment} and Table \ref{table:ab_experiment2}. The results underscore that the Winograd algorithm exhibits more sensitivity to quantization granularity, requiring the finest granularity even with int8 quantization. In contrast, the SFC maintains acceptable accuracy under int8 quantization without specialized quantization. However, at lower bitwidths, it is still necessary to employ frequency-wise quantization for activation tensors.

\begin{table}[h]
\small
  \caption{Ablation study for Resnet18 on int8 quantization. }
  \label{table:ab_experiment}
  \begin{center}
  \begin{tabular}{c c c c }
    \hline
    Algorithm&\thead{Activation \\ Granularity}&\thead{Filter \\ Granularity}&Accuracy\\
    \hline 
   \multirow{4}*{SFC-6(7×7, 3×3)}&Tensor&Channel& 69.18\\
   ~&Freq.&Channel& 69.54\\
   ~&Freq.&Freq.& 69.55\\
   
   ~&Freq.&Channel+Freq.& 69.58\\
   \hline 
   \multirow{2}*{Wino(4×4, 3×3)}&Tensor&Channel& 57.40\\
   ~&Freq.&Channel+Freq.& 67.62\\
    \hline
  \end{tabular}
  \end{center}
\end{table}

\begin{table}[h]
\small
  \caption{Ablation study for Resnet-18 with SFC(7×7, 3×3).}
  \label{table:ab_experiment2}
  \begin{center}
  \begin{tabular}{c c c c }
    \hline
    Quant. Granularity&int8&int6&int4\\
    \hline 
   A: Tensor, W:Channel&69.18&65.42& 17.81\\
   A: Freq., W:Channel&69.54&68.34& 31.10\\
   A: Freq., W:Freq.+Channel&69.58&68.83& 55.82\\
    \hline
  \end{tabular}
  \end{center}
\end{table}

\section{Conclusion}
We propose a novel fast convolution algorithm extended from the Fourier transform, which solves the accuracy problem of the conventional fast convolution algorithm applied to quantized models. According to experiment results, our algorithms outperform state-of-the-art fast convolution quantization works on both model accuracy and computation cost reduction. Our algorithms share exactly the same computational flow as the Winograd algorithm, which means that they can be deployed on general-purpose processors (CPUs, GPUs) and hardware accelerator design conveniently by following the existing works.

\section*{Acknowledgements}
This work was supported in part by the National Key Research and Development Program of China under Grant 2022YFB4400900, in part by the Natural Science Foundation of China under Grants 62211530492 and 62371223.

\section*{Impact Statement}

This paper presents work whose goal is to advance the field of 
Machine Learning. There are many potential societal consequences 
of our work, none which we feel must be specifically highlighted here.

\nocite{langley00}

\bibliography{example_paper}
\bibliographystyle{icml2024}

\newpage
\appendix
\onecolumn
\section{Typical Symbolic Fourier Convolution Algorithms}

For SFC-4($4\times4$, $3\times3$) algorithm, the matrices (with polynomial multiplication integrated) are:
\begin{equation}
\begin{aligned} 
B^T=\begin{pmatrix}
0 & 1  & 1 & 1 & 1 & 0 \\
0 &  -1& 1 & -1& 1 &0\\
0 &  1 & -1 & -1& 1&0\\
0 &  0 & -1 & 0 & 1&0\\
0 &  1 & 0 & -1&0 & 0\\
1 & 0  & 0 & 0 & -1 &0\\
0 & -1 & 0 & 0  & 0 &1\\
\end{pmatrix}, \\
G=\begin{pmatrix}
 1 & 1 & 1\\
 1 & -1& 1\\
 1& -1& -1\\
 1& 0&  -1\\
 0& -1 & 0\\
 1 & 0 & 0 \\
 0&  0 & 1 \\
\end{pmatrix}, 
A = \frac{1}{4}\begin{pmatrix}
 1 & 1 & 1 &1  \\
 1 & -1 & 1 &-1  \\
 0 &2 &0 &-2 \\
 2 &-2& -2& 2\\
 -2& -2& 2& 2\\
 4& 0& 0&0 \\
 0& 0& 0& 4
\end{pmatrix}
\end{aligned}
\end{equation}
It costs 49 multiplications to generate 16 outputs. And only 46 multiplications are consumed when Hermitian symmetry is fully considered.

The SFC-6($6\times6$, $3\times3$) algorithm costs 100/88 multiplications to generate 36 outputs. The matrices are:
\begin{equation} 
\begin{aligned} 
B^T=\begin{bmatrix}
0 & 1  & 1 & 1 & 1 & 1 & 1 &0\\
0 &  1& 1 & 0 & -1 & -1& 0 &0\\
0 &  0 & -1 & -1& 0& 1 & 1 &0\\
0 &  1 & 0 & -1&-1 & 0 & 1 &0\\
0 &  1 & 0 & -1 & 1& 0& -1 &0\\
0 &  0 &-1& 1 & 0 & -1 & 1 &0\\
0 &  1 & -1& 0 &  1& -1& 0 &0\\
0 &  1 & -1& 1 &  1& -1& 1 &0\\
1 & 0  & 0 & 0  & 0&  0& -1&0\\
0 & -1 & 0 & 0  & 0&  0& 0 &1\\

\end{bmatrix}, \\
G=\begin{bmatrix}
 1 & 1 & 1 \\
 0 & 1 & 1\\
-1 & -1& 0\\
-1& 0& 1\\
-1& 0& 1\\
1 & -1 & 0\\
0 & -1& 1 \\
1 & -1&  1\\
1 & 0 &0\\
0&  0 & 1 \\
\end{bmatrix}, 
A = \frac{1}{6}\begin{bmatrix}
 1 & 1 & 1 &1 &1 & 1  \\
  2 & 1 & -1 &-2 &-1 & 1 \\
 -1 & 1 & 2 & 1 & -1 & -2\\
 -1 & -2 & -1 &1 & 2 & 1\\
 1 & -2 & 1 &1 & -2 & 1\\
 1 & 1 & -2 &1 & 1 & -2\\
 -2 & 1 & 1 &-2 & 1 & 1\\
  -1 & 1 & -1 &1 & -1 & 1\\ 
  6& 0   &  0 &0 & 0  & 0\\
  0& 0   &  0 &0 & 0  & 6\\
\end{bmatrix}
\end{aligned}
\end{equation}

The SFC-6($7\times7$, $3\times3$) algorithm costs 144/132 multiplications to generate 49 outputs. Its transformation matrices are:
\begin{equation} \small
\begin{aligned} 
B^T=\begin{bmatrix}
0 & 1 & 1 & 1 & 1 & 1 & 1&0&0\\     
0 & 1 & 1 & 0 & -1& -1& 0&0&0\\    
0 & 0 & -1& -1& 0& 1 & 1&0&0\\    
0 & 1 & 0 & -1& -1& 0 & 1&0&0\\   
0 & 1 & 0 & -1 &  1& 0 &-1&0&0\\    
0 & 0 & -1& 1 &  0& -1& 1&0&0\\
0 & 1 & -1& 0 &  1& -1& 0&0&0\\
0 & 1& -1 & 1& -1 & 1& -1&0&0\\   
1 & 0  & 0 & 0  & 0&  0& -1&0&0\\
0 & -1 & 0 & 0  & 0&  0& 0 &1&0\\
0 & -1 & 0 & 0  & 0&  0& 0 &1&0\\
0 & 0  & -1& 0  & 0&  0& 0 &0&1\\

\end{bmatrix}, \\
G=\begin{bmatrix}
 1 & 1 & 1 \\
 0 & 1 & 1\\
-1 & -1& 0\\
-1& 0& 1\\
-1& 0& 1\\
1 & -1 & 0\\
0 & -1& 1 \\
1 & -1&  1\\
1&0&0\\
0&  0 & 1 \\
 0 &1&0\\
 0 & 0 &1 \\
\end{bmatrix}, 
A = \frac{1}{6}\begin{bmatrix}
 1 & 1  & 1 &1 &1 & 1 &1 \\
  2 & 1 & -1 &-2 &-1 & 1&2 \\
 -1 & 1 & 2 & 1 & -1 & -2&-1\\
 -1 &-2 & -1 &1 & 2 & 1&-1\\
 1 & -2 & 1 &1 & -2 & 1&1\\
 1 & 1  & -2 &1 & 1 & -2&1\\
 -2 & 1 & 1 &-2 & 1 & 1&-2\\
-1 & 1 & -1 &1 & -1 & 1&-1\\ 
  6& 0   &  0 &0 & 0& 0  & 0\\
  0& 0   &  0 &0 & 0& 6  & 0\\
  0& 0   &  0 &0 & 0& 0  & 6\\
  0& 0   &  0 &0 & 0& 0  & 6\\
\end{bmatrix}
\end{aligned}
\end{equation}

The SFC-6($6\times6$, $5\times5$) algorithm costs 196/184 multiplications to generate 36 outputs. Its transformation matrices are:
\begin{equation} \small
\begin{aligned} 
B^T=\begin{bmatrix}

0 &0 & 1 & 1 & 1 & 1 & 1 & 1&0&0\\     
0 &0 & 1 & 1 & 0 & -1& -1& 0&0&0\\    
0 &0 & 0 & -1& -1& 0& 1 & 1&0&0\\    
0 &0 & 1 & 0 & -1& -1& 0 & 1&0&0\\   
0 &0 & 1 & 0 & -1 &  1& 0 &-1&0&0\\    
0 &0 & 0 & -1& 1 &  0& -1& 1&0&0\\
0 &0 & 1 & -1& 0 &  1& -1& 0&0&0\\
0 &0 & 1& -1 & 1& -1 & 1& -1&0&0\\   
1 & 0  & 0 & 0  & 0&  0& -1&0 &0&0\\
0 &1 & 0  & 0 & 0  & 0&  0& -1&0&0\\
0 &1 & 0  & 0 & 0  & 0&  0& -1&0&0\\
0 &0 & -1 & 0 & 0  & 0&  0& 0 &1&0\\
0 &0 & -1 & 0 & 0  & 0&  0& 0 &1&0\\
0 &0 & 0  & -1& 0  & 0&  0& 0 &0&1\\
\end{bmatrix}, \\
G=\begin{bmatrix}
 1 & 1 & 1 &1 & 1 \\
 -1& -1& 0 & 1 & 1 &\\
 1& 0 &-1 & -1& 0\\
 0 &-1& -1& 0& 1\\ 
 0& 1& -1& 0& 1\\
 -1& 0 &1 & -1 & 0\\
-1& 1&  0 & -1& 1 \\
1& -1&1 & -1&  1\\
1&0&0&0&0\\
1&0&0&0&0\\
0&1&0&0&0\\
0&0&0&1&0 \\
0&0&0&0&1 \\
0&0&0&0&1 \\
\end{bmatrix}, 
A = \frac{1}{6}\begin{bmatrix}
 1  & 1 &1 &1 & 1 &1& 1 \\
1 & -1 &-2 &-1 & 1&2& -1 \\
1 & 2 & 1 & -1 & -2&-1& 2\\
-2 & -1 &1 & 2 & 1&-1& -1\\
-2 & 1 &1 & -2 & 1&1& 1\\
1  & -2 &1 & 1 & -2&1& -2\\
1 & 1 &-2 & 1 & 1&-2& 1\\
1 & -1 &1 & -1 & 1&-1& -1\\ 
  6& 0   &  0 &0 & 0& 0  & 0\\
  6& 0   &  0 &0 & 0& 0  & 0\\
  0& 6   &  0 &0 & 0& 0  & 0\\
  0& 0   &  0 &0 & 0& 6  & 0\\
  0& 0   &  0 &0 & 0& 0  & 6\\
  0& 0   &  0 &0 & 0& 0  & 6\\
\end{bmatrix}
\end{aligned}
\end{equation}
\newpage
\section{Applying the SFC Algorithm to Large-size Convolution Kernels}
Large-kernel convolutional neural networks have recently received extensive research attention in recent years, and the kernel sizes ranging from 7×7 to 51×51. The vanilla Fast Fourier Transform (FFT) is an option available. Since large-size convolutions commonly use depth-wise convolutions, this makes the multiplication complexity of the algorithm $n^2logn$, i.e., the computational complexity of the FFT itself will become dominant. 

As a comparison, our algorithm is accomplished using only addition in the transform stage, so applying our algorithm to large-size convolutional kernels is promising. However, our algorithm itself is not applicable to scaling large numbers of transformation points, which can lead to the appearance of higher-order terms further increasing the real multiplication times. We therefore consider an \textbf{iterative convolution} approach to the operation.

Considering a convolution with a 29×29 kernel size and a 26×26 feature map size, we will describe the computational process of iterative convolution. We split the feature map into 5×5 tiles with 6×6 size, while splite the kernel into 6×6 tiles with 5×5 size. Convolution operation will be performed between each feature map tile and each kernel tile, so we use SFC(6×6, 5×5) to speed it up. Note that at this time the partial convolution results of the each tiles are still summed up in the same pattern as the convolution window sliding, which allows us to use the SFC(5×5, 6×6) algorithm to speed up the process as the feature map has been split into 5×5 tiles, and the kernel has been split into 6×6 tiles in the first iteration. The number of multiplications required by above two iterations is the product of the multiplications in the two SFC algorithms, i.e., 132 × 132 = 17,424 multiplications. Our approach reduces the number of multiplications to just \textbf{3\%} of what is required by direct convolution.

We can increass the number of iterations to computing convolution with more larger sizes. Since the SFC algorithm uses only addition for the transformation, it has an easier, efficient and flexible deployment compared to the FFT method. However, when the convolution size is large enough to require 3 or more iterations, the FFT method will appear to be more advantageous in terms of theoretical computational efficiency.



\end{document}